\documentclass{article}

\usepackage{microtype}
\usepackage{graphicx}
\usepackage{subfigure}
\usepackage{booktabs} 

\usepackage{multicol}

\usepackage{hyperref}

\usepackage[accepted]{icml2018}

\begin{document}


\twocolumn[
\icmltitle{Training an Assassin AI for The Resistance: Avalon}




\icmlsetsymbol{equal}{*}

\begin{icmlauthorlist}
\icmlauthor{Robert Chuchro}{equal,to}
\end{icmlauthorlist}

\icmlaffiliation{to}{Department of Computer Science, Stanford University}
\icmlcorrespondingauthor{Robert Chuchro}{chuchro3@cs.stanford.edu}

\vskip 0.3in
]

\printAffiliationsAndNotice{\icmlEqualContribution}




\section{Abstract}
The Resistance: Avalon is a partially observable social deduction game. This area of AI game playing is fairly undeveloped. Implementing an AI for this game involves multiple components specific to each phase as well as role in the game. In this paper, we plan to iteratively develop the required components for each role/phase by first addressing the Assassination phase which can be modeled as a machine learning problem. Using a publicly available dataset from an online version of the game, we train classifiers that emulate an Assassin. After trying various classification techniques, we are able to achieve above average human performance using a simple linear support vector classifier. The eventual goal of this project is to pursue developing an intelligent and complete Avalon player that can play through each phase of the game as any role.

\section{Introduction}

\subsection{Partially Observable Games}

Other than for entertainment, developing AI for games also serves as a good test bed for AI techniques. As it turns out, most real world environments are not completely observable. So, developing techniques in partially observable games that can generalize well can serve a greater purpose than simply mastering specific games.

Traditionally, this has been an extremely difficult problem to tackle even in restricted environments such as games. Only recently have techniques been developed in stochastic and partially observable games such as Poker that achieve super human performance. Our goal with this project is to push the limit further by exploring a more complex game in The Resistance: Avalon.

\subsection{Game Rules}
The Resistance: Avalon is a game where each player is given a role that is only known to them. The roles that we are focused on for this project are Merlin, Percival, Loyal Servant, Assassin, and Morgana. See figure \ref{role_descrip} for a description of the roles. In this analysis we will focus primarily on 5 player games, which contain one of each of the aforementioned roles.

The first phase of the game involves each player taking turns nominating a specified number of players to go on a mission. Then each player votes to approve the nomination or not. If a majority of players approve the team, the team members do the mission and each are given a succeed and fail card. These cards are handed face-down to the mission leader, who shuffles them face-down so that the outcome is anonymous. They are then revealed to see if the mission succeeded or not. A round succeeds if and only if there are no fail cards played. The next mission team will be proposed by the person clockwise to the first team leader.

If a majority of players reject the team, or it is a tie, the members do not do the mission. Instead, leadership passes clockwise, and the new leader proposes a new team. The proposal is voted on again, and the process repeats until a team is agreed upon. Note that if five teams in a single round are rejected in a row, the evil team automatically wins the entire game. This incentivizes everyone to approve the 5th proposal of a round.

The goal for the resistance team is to succeed 3 missions out of up to 5 missions. If 3 missions fail, the game is over and the spy team wins. If 3 missions succeed, the game proceeds to the assassination phase, where the assassin attempts to identify the resistance player with the Merlin role by selecting a player to be shot. If the assassin is correct, the spies win. If the assassin shoots any player but Merlin, the resistance have won.

\begin{figure}[h]
\begin{itemize}
    \item Merlin: A member of the Resistance who knows all spy members.
    \item Percival: A member of the Resistance who knows of Merlin(s).
    \item Loyal Servant: A member of the Resistance with no additional information.
    \item Assassin: A member of the Spies who can attempt to shoot Merlin to win the game after the Mission phase.
    \item Morgana: A member of the Spies who appears as Merlin to Percival.
\end{itemize}
  \caption{Description of Roles}
  \label{role_descrip}
\end{figure}

\section{Related Work}
\subsection{The Resistance AI competition}
In 2012, Dagstuhl Seminar 12191 held a tournament for the original game The Resistance \cite{lucas_et_al:DR:2012:3651},. In this tournament, each team submitted one player that would be evaluated on up to 10000 matches. They measured performance based on winrate against the other AIs. The report states that simple strategies seem to outperform complex strategies over the course of many games \cite{lucas_et_al:DR:2012:3651}.

One of the key differences between this tournament on our goal is that the original Resistance game only had two basic roles: a standard resistance member, and standard spy. Adding unique roles adds a lot of complexity to the game, but the outcomes of this seminar can serve as a base to build our AI from.

\subsection{Finding Friend and Foe in Multi-Agent Games}
Very recently, Serrino et. al. designed an agent named DeepRole to play Avalon online \cite{deeprole}. They built a fully functional player using only general learning techniques such as Deep Reinforcement Learning and Monte Carlo methods. They claim the player does not have any game knowledge built in. A lot of the techniques they used follow from a recently developed Poker player named DeepStack. Both of these games contain information that is hidden and specific to certain players, while other information is public to everyone. 

In the last couple years, Poker AIs have begun to reach super human performance in setting such as no limit Texas Hold'em, and the authors believe Avalon to be a more difficult problem \cite{deeprole}. In our pursuit of developing an AI for the game Avalon, we hope to combine various general learning techniques, but some simple feature engineering can be done to incorporate game knowledge in order to make the AI more competitive.

\section{Concerns}

One major concern with creating an AI for Avalon is that the game contains a lot of dialog as players try to communicate their perspective and offer information. The dataset that we have does not contain any saved dialog that occurred in the game, and even if it did, this becomes a very difficult NLP problem to try to extract relevant chat.

\section{Data}

There currently exists an online version of this game, which offers anonymized game data from every game played on the site. There are over 10000 recorded game histories available for download as a JSON file. The data is available on \url{http://proavalon.com/statistics}. There are almost 6000 game histories from 5 player games which offers sufficient data for training a player through game history. 

The information we plan to use from each game is the table of mission proposals and vote history per player, as well as outcomes of each mission. Finally, we will also use the outcome of the assassination phase.

\subsection{Features}

In the initial attempt, we tried to generate a large feature vector that does not incorporate any game specific knowledge. For each game, the features are a 5x5x5x4 matrix. This represents a length 4 vector for each potential player (5) by each potential sub-mission (5) by each potential mission (5). The length 4 vector contains whether for sub-mission $j$ of mission $i$, player $p$ selected the mission, was selected on the mission, their vote, and whether they are a member of the resistance or not. Each value was 1 if true, or -1 if false. In addition, the value is 0 if this mission never happened (i.e game ended after mission 4 so there was no mission 5). Since our learning algorithms expect a consistently sized input vector, it was essential to include this in our modeling of the feature space.

\subsection{Feature Engineering}

We initially took a very greedy approach with how vague our feature space is, and evidently, there is not enough data for our algorithms to be able to train effectively. After experimenting with our proposed methods on this general dataset, the next step was to reconstruct the feature vector by incorporating game-specific knowledge to enrich the features. 

We incorporated game-specific knowledge to create a more compact feature space. We generated a set of 9 features based on our understanding of important information from the vote history. In order to reduce the feature space to as small as possible, we trained a simple linear support vector classifier for each set in the powerset of our 9 features and kept the feature set that had the highest accuracy. Since training a linear classifier took only seconds, we are able to iterate the powerset of features in a reasonable amount of time. 

After iterating through our powerset iteration algorithm we reduced our feature space to 4 numbers for each role. For each player (5), we narrowed down that the following information was key:
    
    \begin{itemize}
        \item the number of correct votes (approving clean or rejecting dirty)
        \item the number of correct proposals
        \item whether the player was the first to propose the entirely correct team (0 or 1)
        \item whether their first pick was correct or not (0 or 1)
    \end{itemize}
    
Note that if that role is a spy, the entire feature vector is set to all values of 0. This prevented our classifiers from selecting a spy as Merlin. In addition, we index each player in our features corresponding to the player with the first pick of the game having the first index of the feature set.

\section{Evaluation}

\subsection{Training an Assassin}

The initial phase we would like to tackle is the Assassination phase of the game since it is a much more straight forward problem. This problem can be modeled as follows: Given the mission and vote history of a game in which 3 missions succeeded, predict which member of the Resistance is Merlin.

\subsection{Baseline}

The initial baseline to this problem is a random Assassin that simply shoots a Resistance member uniformly at random. In a 5 player game, there are 3 Resistance members, so the probability of being correct in expectation is $\frac{1}{3}$.

\subsection{Oracle}

A trivial oracle in this situation is simple: An oracle Assassin would always shoot Merlin correctly 100\% of the time. In order to gain more insight to this problem,  we will compare ourselves to human performance which is available from the dataset.

\subsection{Goal Results}

One goal for the Assassin AI would be to outperform the average human player that plays online from our data source. Preliminary analysis of the game data indicates that when the Resistance succeeds 3 missions in 5 player games, the Assassin shoots correctly 44.8\% of the time (Figure \ref{baseline_accuracy_results}).

One way to measure performance is by training our player on a subset of the data (between 80-90\%), then testing it on the remaining data (10-20\%).

We measure performance accuracy using K-fold Cross Validation with 10 folds \cite{cross_validation}. We randomize the dataset, then split the data into 10 equal sized sets. For each set, we train on the remaining 90\% of the data, then use the set as a validation set and measure the prediction accuracy on this set. We report the average of the 10 accuracies as the accuracy in our results.

\section{Methods}

\subsection{Supervised Learning}

Since our training data is labeled with what the correct choice is for every game, we can model our problem as a supervised learning problem. This can be tackled with approaches we already touched on in class such as Linear Classifiers or Approximate Q-Learning.

As mentioned in the data section, our input for each game is the the role of each player, vote history (mission proposals and each player's corresponding vote), and the mission outcomes (success/fail). The correct label on each
 game is the player who is actually Merlin and should have been assassinated. This label can be used as a reward or identifying mislabels.
 
\subsection{Deep Learning}

Since a linear functions cannot model very complex data, using multilayer perceptrons, or a neural network, can offer an improvement to the modeling power of our Assassin. With several thousands games to analyze, this should be enough data to train a neural network.

For out implementation of a neural network classifier, we will use Python with the Tensor Flow library \cite{tensorflow}.

\subsection{Support Vector Classifier}

To start, we will implement a multi class linear support vector machine and use this as a classifier to predict the most likely resistance member to be Merlin. Since support vector machines typically classify between two classes, we will use the one vs. rest technique. That is, we will train one support vector machine for each class that attempts to classify the input as that class or not that class. So for a 5 player game, we will need 5 classifiers as described. For each input, each classifier outputs a confidence score, and we take the maximum over these 5 scores and classify that player as Merlin.

For our implementation, we are using Python with Scikit-Learn's implementation of a Support Vector Machine \cite{scikit-learn}. We trained our classifier using scikit-learn's LinearSVC algorithm over 100000 iterations using squared hinge loss and an L2 penalty. In addition, we later experiment with using non-linear support vector classifiers such as a Radial Basis Function (RBF) Kernel.

\section{Results}

\subsection{Deep Learning Results}

After experimenting with different sized fully connected networks ranging from widths as small as 8 to as large as 256, we reported our accuracy for the best sizes. When using our large general feature set, we were able to achieve the highest training accuracy, but also the lowest test accuracy (table \ref{general_accuracy_results}). This is a clear sign of overfitting the training data. When using the engineered feature set which is considerably smaller, we were able to reduce overfitting, achieving a higher test accuracy, but correspondingly lower training accuracy (Figure \ref{feature_accuracy_results}).

\begin{figure}[h]
  \centering
  \includegraphics[width=0.5\textwidth]{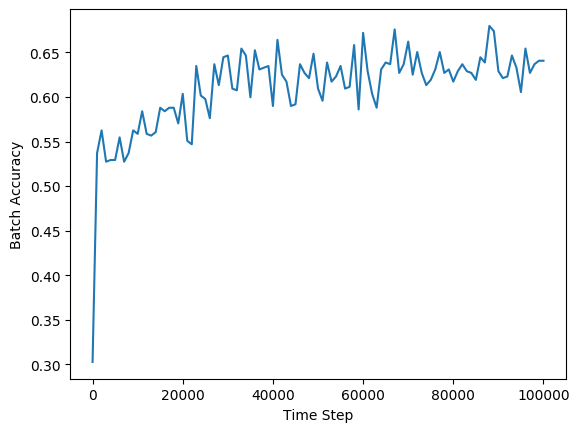}
  \caption{Neural Network Training Accuracy}
  \label{nn_train_acc}
\end{figure}

\begin{figure}[h]
  \centering
  \includegraphics[width=0.5\textwidth]{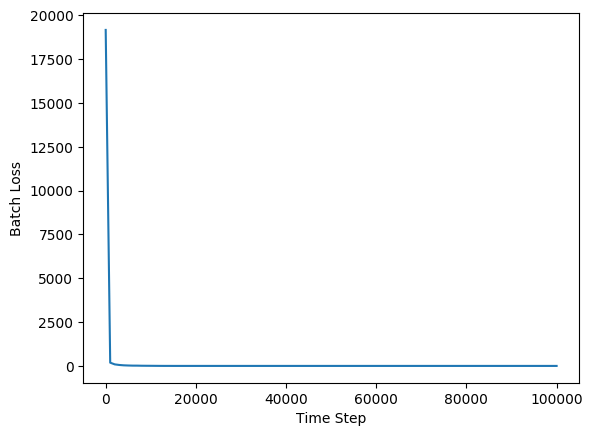}
  \caption{Neural Network Training Loss}
  \label{nn_train_loss}
\end{figure}

In addition, the neural network produced very inconsistent results, even when using small starting learning rates ($< 1e^{-5}$) combined with the Adam Optimizer \cite{adam_optimizer}. Even though our loss was consistently able to converge extremely quickly (Figure \ref{nn_train_loss}), training accuracy continued to be unstable as seen in Figure \ref{nn_train_acc}. The result was also extremely inconsistent test accuracies, so it was difficult to gauge overall performance. However, even the top models did not surpass average human performance on the validation set.

 \begin{figure}[h]

    \caption{Neural Network Architecture}
    \begin{tabular}{ |c||c|p{1.0cm}|c|  }
 \hline
 Layer & Input & output & Activation\\
 \hline
 \hline 
 Fully Connected-1 & Features & FC-2& ReLU \\
 \hline
  Fully Connected-2 & FC-1 & FC-3 & ReLU \\
 \hline
 Fully Connected-3 & FC-2 & Scores  & Softmax \\
 \hline
\end{tabular}
\label{nn-architecture}
\end{figure}

\begin{figure}[h!]
  \caption{Baseline Accuracy Results}
    \begin{tabular}{ | p{4.cm} || p{1.4cm} | p{1.4cm} |}  
    \hline
    Algorithm & Training Accuracy & Test Accuracy\\ \hline
    Random & N/A & 0.333 \\ \hline
    Human & N/A & 0.448 \\
    \hline
    \end{tabular}
  \label{baseline_accuracy_results}
\end{figure}

\subsection{Support Vector Classifier Results}

When training a linear support vector classifier on the general feature set, the performance was better than the neural network classifier (Figure \ref{general_accuracy_results}). However, the model presented similar issues with overfitting the training data, just not as extreme. 

When we trained the same linear support vector classifier on our engineered feature set, we achieved the overall top performing classifier (Figure \ref{feature_accuracy_results}). This was the only classifier that outperformed the average human on the validation set by 1.1\%. 

In addition, we experimented with using non-linear support vector classifiers. In particular, train an non-linear Radial Basis Function (RBF) kernel on the engineered feature set. While the non-linear RBF was able to achieve a modest test accuracy of .419 (Figure \ref{feature_accuracy_results}), it also showed signs of over-fitting when compared to the linear model (55.7\% vs 47.5\% training accuracy).

\begin{figure}[h!]
  \caption{Accuracy Results using general vote history features}
    \begin{tabular}{ | p{4cm} || p{1.4cm} | p{1.4cm} |}  
    \hline
    Algorithm & Training Accuracy & Test Accuracy\\ \hline
    Linear Support Vector Classifier & .603 & 0.372 \\ \hline
    256x128x128 NN with 256 batch size & .642 & 0.342 \\
    \hline
    \end{tabular}
  \label{general_accuracy_results}
\end{figure}

\begin{figure}[h!]
  \caption{Accuracy Results using engineered features}
    \begin{tabular}{ | p{4cm} || p{1.4cm} | p{1.4cm} |}  
    \hline
    Algorithm & Training Accuracy & Test Accuracy\\ \hline
    \textbf{Linear Support Vector Classifier} & \textbf{.475} & \textbf{0.459} \\ \hline
    16x16x8 NN with 256 batch size & .496 & 0.405 +/- 0.043 \\
    \hline
        Non-Linear RBF Support Vector Classifier & .557 & 0.419 \\ \hline
    \end{tabular}
  \label{feature_accuracy_results}
\end{figure}

\section{Error Analysis}

\begin{figure}[h]
  \centering
  \includegraphics[width=0.48\textwidth]{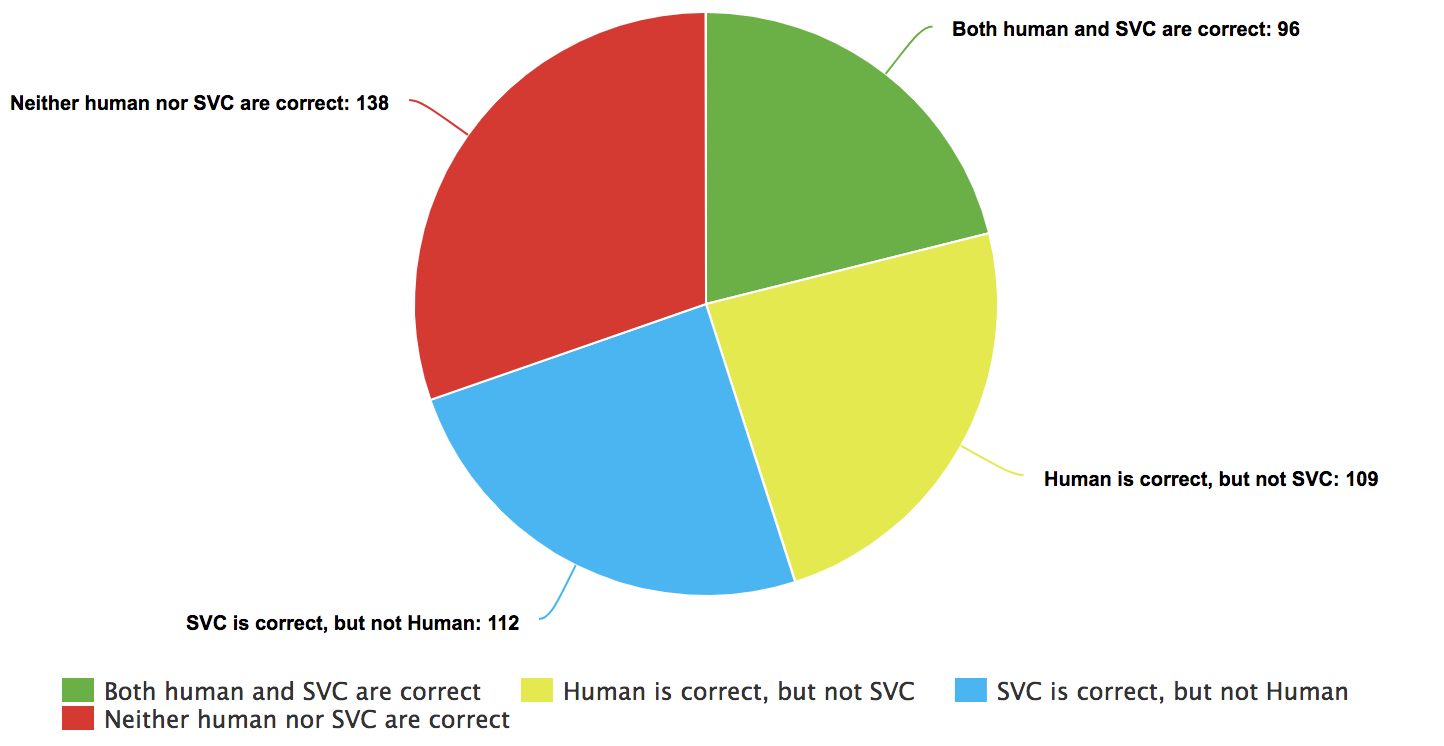}
  \caption{Breakdown of Human vs SVC Correct Classification Counts}
  \label{error_pie}
\end{figure}

In an attempt to gain insight into the misclassification by our top performing classifier, we start by comparing each classification with the classification of the human assassin. Interestingly, there seems to be very little correlation between the games that the human assassinates correctly and the games the linear classifier assassinates correctly (Figure \ref{error_pie}). The counts for our linear SVC vs human classification on one of our validation sets of 455 games are as follows:

\begin{itemize}
    \item  Both human and SVC are correct: 96
\item Human is correct, but not SVC: 109
\item SVC is correct, but not Human: 112
\item Neither human nor SVC are correct: 138
\end{itemize}

\begin{figure}[h]
  \centering
  \includegraphics[width=0.48\textwidth]{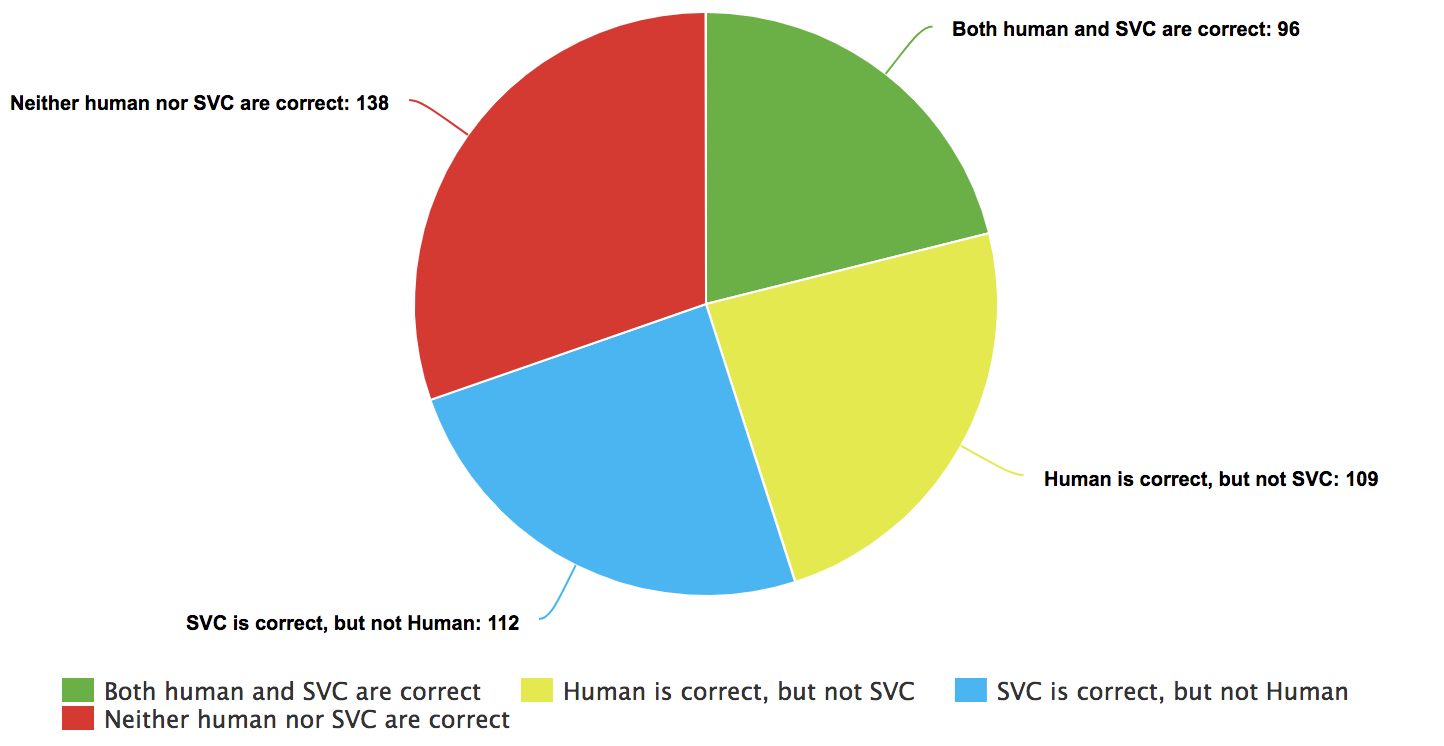}
  \caption{Breakdown of which role is shot}
  \label{shot_breakdown}
\end{figure}

In addition, when classifying incorrectly, the linear SVC shoots Percival 52.2\% of the time, while shooting the Loyal Servant 47.8\% of the time. We hypothesized that Percival would be shot significantly more often than the Loyal Servant when shooting incorrectly, but surprisingly, this did not turn out to be the case with only a slight bias towards Percival.

\section{Conclusion}

A Linear Support Vector Classifier was able to consistently achieve the highest test performance using our smaller engineered features. It even surpassed the average human performance. Although human performance was similar to Linear SVC, there was almost no correlation between which games the human and classifier shot correctly.

As it turns out, experimenting with more complex models only led to over-training. When using non-linear support vector classifiers or neural networks, our training accuracy would be consistently higher, but our test accuracy was lower. In addition, the neural network did not achieve consistent results despite converging to a small loss very quickly. Our results follow the philosophy of Occam's Razor: a simple model generalizes better than a complex one with more variables.

\section{Future Work}

A next step would be to continue training our support vector classifier on games sizes larger than 5 to see how well our feature set generalizes. It is interesting to note that human assassination performance remains about the same (43-44\%) even as the number of resistance players increases. We predict that classification will get increasingly difficult as the number of players/classes increases for our classifier; so, continuing to surpass average human performance will be a challenge.

Ultimately, we would like a fully functional player for the entire game, not just the Assassination portion. Serrino et. al. were able to train a fully functional and competitive player to play 5 player Merlin and Assassin games using general learning techniques \cite{deeprole}. We believe that by incorporating game knowledge combined with general learning techniques, we can build a complete player that can achieve super human performance.

\bibliographystyle{icml2018}
\bibliography{citations}

\end{document}